\documentclass[10pt,twocolumn,letterpaper]{article}

\usepackage{cvpr}              
\usepackage{float}
\definecolor{cvprblue}{rgb}{0.21,0.49,0.74}

\usepackage[pagebackref,breaklinks,colorlinks,allcolors=cvprblue]{hyperref}

\def\paperID{4221} 
\def\confName{CVPR}
\def\confYear{2025}

\title{MAR-3D: Progressive Masked Auto-regressor for  \\ High-Resolution 3D Generation}

\author{Jinnan Chen$^{1}$ \qquad Lingting Zhu$^{2}$ \qquad Zeyu Hu$^{3}$ \qquad Shengju Qian$^{3}$  \\ 
\qquad Yugang Chen$^{3}$  \qquad Xin Wang$^{3}$  \qquad Gim Hee Lee$^{1}$ \vspace{1mm} \\
$^{1}$National University of Singapore \\
$^{2}$The University of Hong Kong\\
$^{3}$LIGHTSPEED\\
\multicolumn{1}{c}{\textcolor{orange}{\tt \href{https://jinnan-chen.github.io/projects/MAR-3D/}{\textbf{https://jinnan-chen.github.io/projects/MAR-3D/}}}}
}

\begin{document}
\twocolumn[{
    \maketitle
\begin{center}\centering\includegraphics[width=0.95\textwidth]{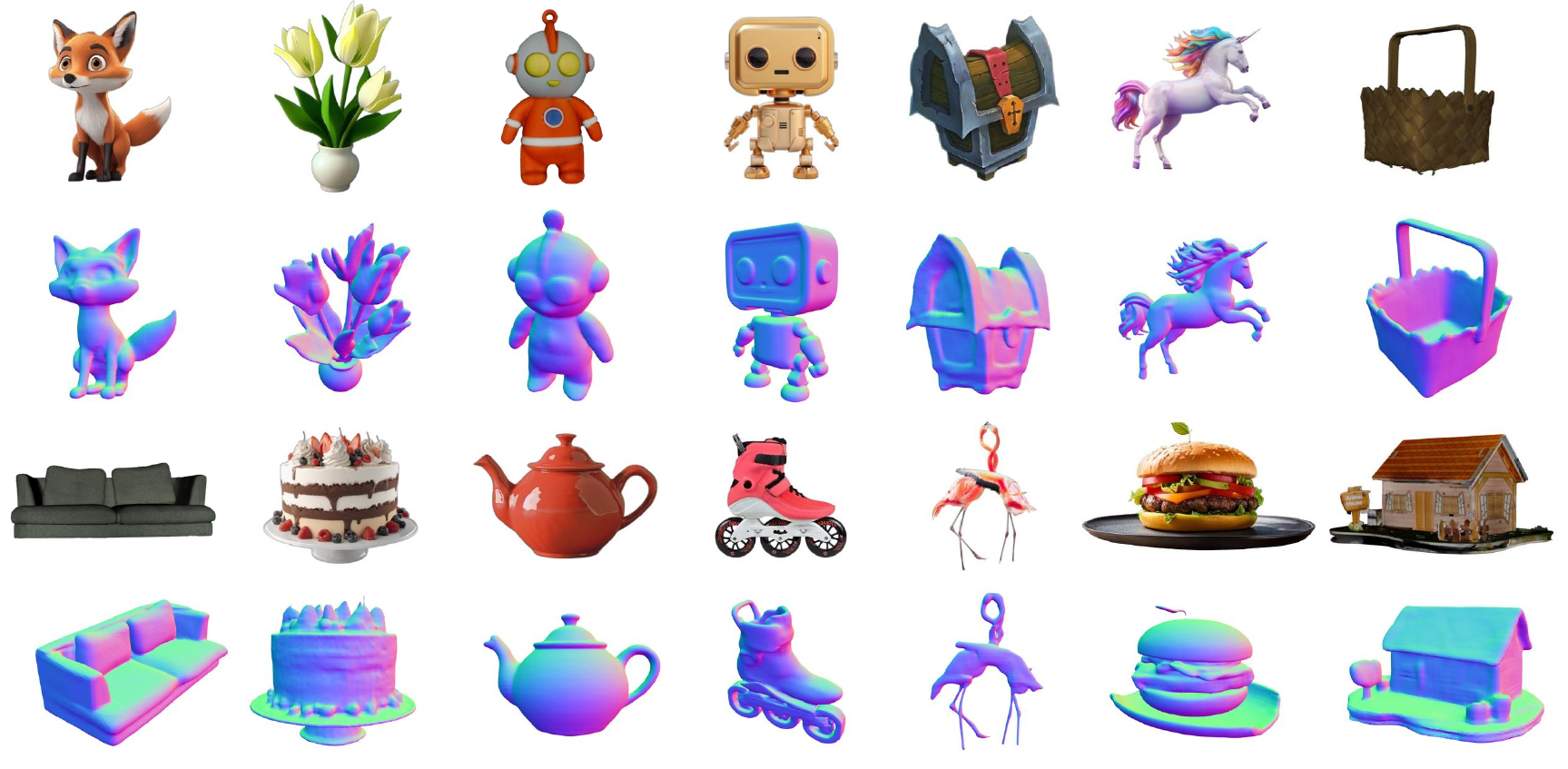} \captionsetup{type=figure}
        \caption{Our MAR-3D demonstrates 
        strong generalization ability on in the wild images, accurately handling complex geometric details including fine structures and intricate shapes. We visualize the normal maps of our generated meshes with some random rotations.} 
        \label{fig:teaser}
    \end{center}
}]
\begin{abstract}
Recent advances in auto-regressive transformers have revolutionized generative modeling across different domains, from language processing to visual generation, demonstrating remarkable capabilities. However, applying these advances to 3D generation presents three key challenges: the unordered nature of 3D data conflicts with sequential next-token prediction paradigm, conventional vector quantization approaches incur substantial compression loss when applied to 3D meshes, and the lack of efficient scaling strategies for higher resolution latent prediction. To address these challenges, we introduce MAR-3D, which integrates a pyramid variational autoencoder with a cascaded masked auto-regressive transformer (Cascaded MAR) for progressive latent upscaling in the continuous space. Our architecture employs random masking during training and auto-regressive denoising in random order during inference, naturally accommodating the unordered property of 3D latent tokens. Additionally, we propose a cascaded training strategy with condition augmentation that enables efficiently up-scale the latent token resolution with fast convergence. Extensive experiments demonstrate that MAR-3D not only achieves superior performance and generalization capabilities compared to existing methods but also exhibits enhanced scaling capabilities compared to joint 
distribution modeling approaches (e.g., diffusion transformers).
\end{abstract}    
\section{Introduction}
\label{sec:intro}
High-quality 3D mesh generation has emerged as a critical challenge in computer graphics and vision, with widespread applications in gaming and AR/VR industries. Recent advances in open-world 3D object generation have demonstrated promising results following several distinct paradigms. Large Reconstruction Models~\cite{lrm24,TripoSR2024} employ transformers to convert single images into 3D shapes through implicit representations with multi-view supervision. However, the lack of generative priors leads to blurry artifacts in unseen regions. Another direction combines 2D diffusion models~\cite{imagedream23, mvdream23, liu2023syncdreamer, hgm, long2023wonder3d} with sparse view 3D reconstruction models~\cite{lgm24,instant3d,xu2024instantmesh,grm24}. However, these methods are constrained by the quality and consistency of the generated multi-view images. A more recent paradigm follows the success of 2D image generation by utilizing 3D variational auto-encoders (VAE) to compress dense point clouds into latent space, then applying diffusion models for direct 3D shape generation~\cite{3dshape23,michelangelo23,craftsman24,zhang2024clay,hu2024xray}.

While these approaches show promising performance, they face fundamental challenges in effectively increasing tokens to achieve higher quality generation: (1) existing VAEs and generators struggle to maintain geometric detail when representing 3D data into limited number of tokens, (2) the computational complexity of transformer-based generators grows quadratically with token resolution, directly increasing the number of tokens requires hundreds of GPUs~\cite{zhang2024clay}—making it impractical,  and (3) the lack of an effective strategy to progressively increase resolution while maintaining generation quality.

In this work, we address these challenges through a progressive approach. First, we introduce a Pyramid VAE that captures multi-scale geometric information through different cross-attention layers, improving data representation and reconstruction quality while maintaining efficient token resolution. Second, we develop a cascaded generation strategy using two Masked Auto-Regressive (MAR) models: MAR-LR generates low-resolution tokens that capture overall shape, while MAR-HR refines these into high-resolution tokens with fine geometric details.

This progressive strategy is enabled by several key strategies: (1) a random masking operation during training that naturally fits the unordered property of 3D latent tokens, (2) condition augmentation that reduces compounding errors when increasing resolution, and (3) an efficient parallel decoding strategy that maintains generation quality even with longer token sequences. Through extensive experiments, we demonstrate that our approach not only achieves superior performance compared to existing methods but also exhibits enhanced progressive properties in terms of both latent token resolution and generation quality.

Our key contributions include:
\begin{itemize}
\item A Pyramid VAE architecture that enables effectively increasing token resolution to preserve geometric details.
\item A cascaded MAR generation that combines MAR-LR and MAR-HR models for progressive token generation.
\item A comprehensive study demonstrating the benefits of our progressive up-scaling strategy combining auto-regressive model for improving generation quality.
\item The state-of-the-art results among open-sourced methods on public benchmarks, with particular improvements in preserving fine geometric details and complex structures.
\end{itemize}

\section{Related works}
\label{sec:related}
\subsection{3D Large Reconstruction Models}
The Large Reconstruction Model (LRM)~\cite{lrm24,wei2024meshlrm,TripoSR2024,xie2024lrmzero} marks a pivotal advancement in single-view 3D reconstruction by significantly scaling up both model architecture and dataset size for neural radiance field~\cite{mildenhall2020nerf} (NeRF) prediction. While LRM was initially developed for reconstruction tasks, its capabilities have been extended to text-to-3D and image-to-3D generation through integration with multiview diffusion models~\cite{imagedream23,wang2024crm,shi2023zero123plus,long2023wonder3d}, as demonstrated by subsequent works such as Instant3D~\cite{instant3d}, DMV3D~\cite{xu2023dmv3d}, and InstantMesh~\cite{xu2024instantmesh}.
Recent innovations, including LGM~\cite{lgm24}, GRM~\cite{grm24}, and LaRa~\cite{chen2025lara}, have further enhanced rendering quality and computational efficiency by combining novel view generation diffusion models with generalizable Gaussians family~\cite{kerbl3Dgaussians, huang20242d} in a feed-forward manner. However, this approach of leveraging pre-trained multiview diffusion models~\cite{mvdream23, imagedream23} encounters two significant challenges. First, the disjoint training of reconstruction and diffusion models can introduce multiview inconsistencies during inference. Second, the multiview diffusion process typically produces lower resolution outputs, resulting in the loss of original image details and degraded reconstruction fidelity.
\subsection{3D Generative Models}
Direct generation of 3D content under explicit 3D supervision offers a more efficient approach to 3D content creation~\cite{jun2023shapegeneratingconditional3d,3dshape23,michelangelo23,zhang2024clay,ren2024xcube,surfd,zhu2025muma}. However, training generative models directly on 3D data poses significant challenges due to extensive memory and computational requirements. Recent approaches address these limitations by first compressing 3D shapes into compact latent representations before applying diffusion or auto-regressive models.
This field saw significant advancement with 3DShape2VecSet~\cite{3dshape23}, which introduced a 3D mesh VAE that encodes point clouds into shape latents and decodes occupancy values using grid embeddings, coupled with a categorical latent diffusion model. Building on this foundation, subsequent works including Michelangelo~\cite{michelangelo23}, CLAY~\cite{zhang2024clay}, and Craftsman~\cite{craftsman24} scaled up 3D diffusion models by leveraging diffusion transformer architectures and larger datasets, achieving superior generalization capabilities. Xcubes~\cite{ren2024xcube} further advanced the field by compressing 3D meshes into sparse latent voxels and learning hierarchical voxel diffusion models for generation.
While recent attempts aiming at synthesizing meshes directly~\cite{nash2020polygen,meshgpt23, meshanything24,chen2024meshxl} have demonstrated promising results in high-quality mesh generation, their effectiveness is limited by mesh tokenization constraints and strict requirements on face counts, which ultimately restrict their generalization capability. Distinct from previous approaches, our method preserves the advantages of auto-regressive models while circumventing the limitations of lossy tokenization from vector quantization through the integration of diffusion models and progressive up-scaling strategy.
\subsection{Generative Auto-Regressive Models}
Auto-regressive models~\cite{vaswani2023attentionneed,MaskedAutoencoders2021} have revolutionized visual generation~\cite{vqgan20,nichol2022pointe,Liu2023MeshDiffusion,chang2022maskgit,muse23,var24,llamagen24,mar24} through their sequential approach of synthesizing images using discrete tokens produced by image tokenizers. Pioneering works such as VQGAN~\cite{vqgan20} demonstrated the effectiveness of raster-scan sequences for next-token prediction by first training a discrete-valued tokenizer on images, utilizing a finite vocabulary obtained through vector quantization, followed by per-token prediction. Subsequently, Maskgit and MUSE~\cite{chang2022maskgit,muse23} advanced beyond sequential token prediction by introducing parallel prediction of multiple tokens in random order, substantially improving both generation quality and efficiency.
More recently, VAR~\cite{var24} introduced a novel next-scale prediction paradigm that not only better preserves spatial locality but also achieves significant computational efficiency gains. However, these methods still suffer from information loss due to quantization in the latent space. To address this limitation in continuous domains, MAR~\cite{mar24} proposed modeling per-token probability distributions using a diffusion process~\cite{diffusion20}, effectively combining the efficiency of auto-regressive models with the advantages of continuous diffusion processes to mitigate quantization losses. Building upon these advancements and the demonstrated scalability benefits in 2D applications, our work extends continuous auto-regressive modeling into the realm of high quality 3D mesh generation.
\begin{figure*}[t] 
\centering
\includegraphics[width=0.95\textwidth]{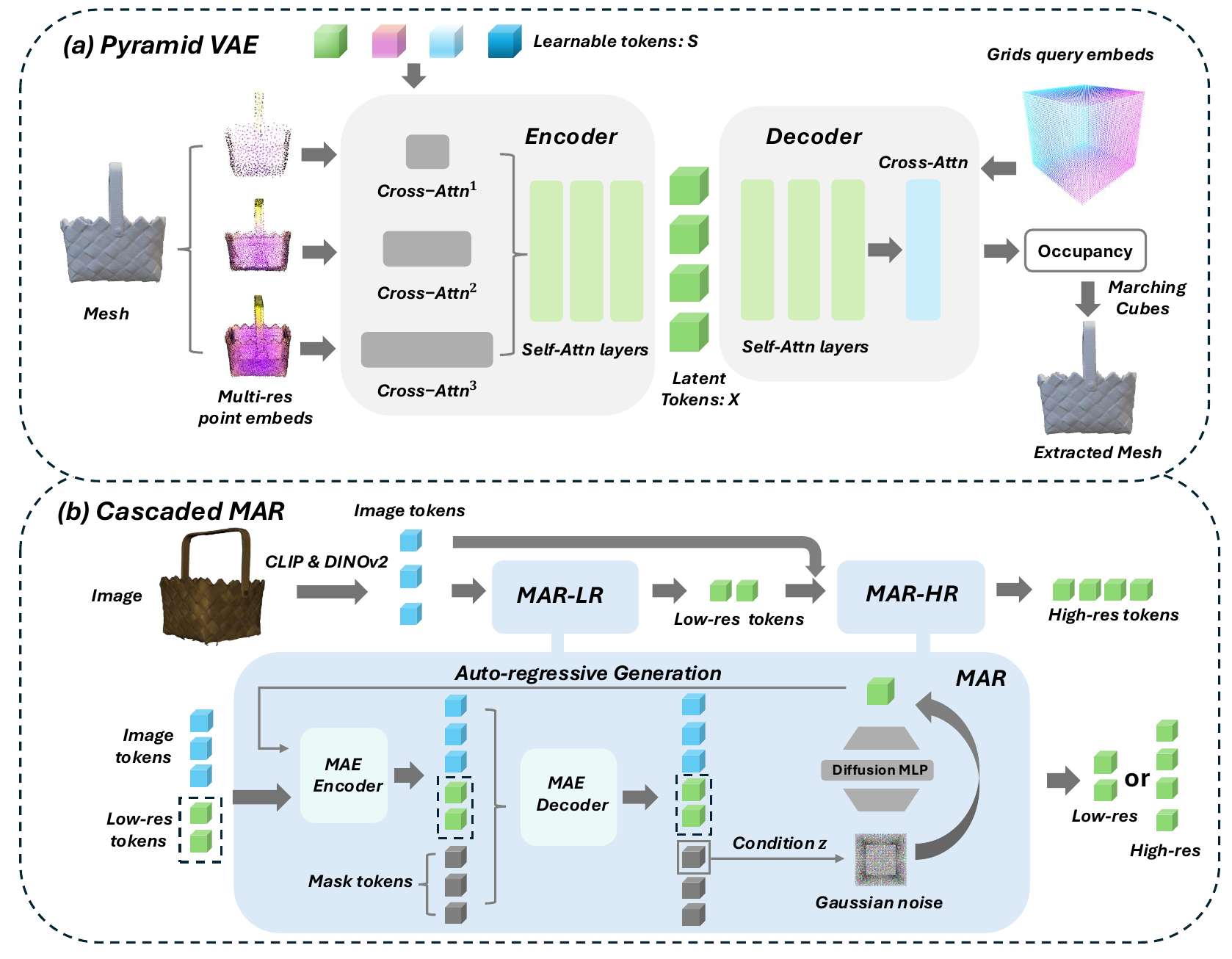} 
\caption{\textbf{Overview of MAR-3D:} (a) Pyramid VAE: It processes learnable tokens through separate cross-attention layers, taking multi-resolution point clouds and normals as input to generate occupancy fields. (b) Cascaded MAR: Conditioned on image features extracted by CLIP and DINOv2, we employ a cascaded design: a MAR-LR model for generating low-resolution tokens, and a MAR-HR model for high-resolution token. The MAR architecture details are illustrated in the blue box. While MAR-LR and MAR-HR share the same architecture, they differ in the inputs: MAR-HR additionally requires low-resolution tokens as input (shown in the dashed box).}
\label{fig:framework}
\end{figure*}
\section{Approach}
\label{sec:method}
\subsection{Overall}
As illustrated in Fig.~\ref{fig:framework}, our framework MAR-3D consists of a Pyramid VAE architecture coupled with a Cascaded MAR. Specifically, MAR-LR generates low-resolution tokens encoding global structure conditioned on the input image tokens, while MAR-HR produces high-resolution tokens conditioned on the previously generated low-resolution tokens and image tokens, enabling fine geometric detail refinement. The final 3D mesh is extracted by applying Marching Cubes to the occupancy field \cite{ocnn} from the high-resolution latent tokens.
\subsection{Pyramid VAE}
\paragraph{Encoder} As illustrated in Fig.~\ref{fig:framework} (a), our Pyramid VAE first generates $K$ levels of down-sampled resolutions from the input point cloud $\mathbf{P}^k \in \mathbb{R}^{N^k \times 3}$, where $N^k$ is the number of points for $k_{th}$ level, and concatenates them with their corresponding surface normals $\mathbf{P}_n^k$ as multi-resolution point embeddings:
\begin{equation}
\hat{\mathbf{P}}^k = [\gamma(\mathbf{P}^k), \mathbf{P}_n^k],
\end{equation}
where $\gamma(\cdot)$ denotes the positional embedding function. Subsequently, we apply cross-attention operations between learnable queries $\mathbf{S}$ and each level of point embedding $\hat{\mathbf{P}}^k$, which serve as keys and values. Each resolution level employs distinct cross-attention layers: coarse levels capture structural features while fine levels extract detailed geometric information, which are then added together. Multiple self-attention layers are subsequently applied to obtain the latent tokens $\mathbf{X}$:
\begin{equation}
\mathbf{X} = \text{SelfAttn}\left(\sum_{k=1}^K \text{CrossAttn}^k(\mathbf{S}, \hat{\mathbf{P}}^k)\right).
\end{equation}
While directly using high-resolution latent tokens would increase computational burden and complicate diffusion training, our hierarchical design enables efficient token length compression while preserving  geometric details. Unlike ~\cite{zhang2024clay}, which randomly operates on one of multiple resolutions of point clouds during training, our Pyramid VAE processes multiple resolution levels simultaneously and also supports latent tokens of different resolutions, which enables progressive scaling in the next step.
\paragraph{Decoder}
The decoder architecture first processes the encoded features through multiple self-attention layers. We then sample query points with positional embeddings and employ cross-attention between these embedded points and the processed latent tokens to predict occupancy values. 
\paragraph{Training Objective}
We optimize the VAE using a combination of binary cross-entropy (BCE) loss for occupancy prediction and KL-divergence loss for latent space regularization:
\begin{equation}
\mathcal{L}_{\text{vae}} = \mathbb{E}_{x \in \mathbb{R}^3}\left[\text{BCE}\left(\hat{\mathcal{O}}(x), \mathcal{D}(\gamma(x), \mathbf{X})\right)\right] + \lambda_{\text{kl}}\mathcal{L}_{\text{kl}},
\end{equation}
where $\hat{\mathcal{O}}(x)$ is the ground truth occupancy value.
\subsection{Cascaded Masked Auto-Regressive Model}
Our Cascaded MAR consists of MAR-LR and MAR-HR models with the same architecture while different input tokens. MAR-LR takes image tokens as condition and MAR-HR takes both of the image tokens and the low-resolution tokens generated by MAR-LR. Traditional auto-regressive models predict tokens sequentially based on previous tokens, following a causal ordering paradigm widely adopted in language models. However, given the unordered nature of our latent tokens, 
we employ a random and parallel decoding strategy inspired by image generation techniques~\cite{chang2022maskgit} to efficiently generate large number of tokens:
\begin{equation}
\begin{aligned}
p(x^1,\ldots, x^N) &= p(X^1, \ldots, X^S) \\
&= \prod_{S} p(X^s | X^1, \ldots, X^{s-1}),
\end{aligned}
\label{eq:marginal}
\end{equation}
where $\mathbf{X}^s$ represents a set of tokens generated in parallel at step $s$, the generation order of latent tokens is determined randomly. During training, we adopt Masked Autoencoders (MAE)~\cite{MaskedAutoencoders2021}, which employs random masking to reconstruct masked regions using information from unmasked tokens. This enables the model to predict tokens in arbitrary order during inference. Since our latent tokens exist in continuous space, we apply diffusion loss~\cite{mar24} rather than conventional Cross Entropy loss to supervise both of our auto-regressive model and diffusion model. During inference, starting from image tokens, the newly generated tokens are positioned in their designated locations and iteratively fed back into the MAE encoder-decoder and diffusion denoising pipeline for subsequent continuous token generation. The low-resolution tokens generated by MAR-LR are concatenated with image tokens and fed into MAR-HR, following the same process to generate final high-resolution tokens, which are fed into our VAE decoder to generate the occupancy field and extract the mesh.
\subsubsection{MAR-LR}
\paragraph{Positional Embedding}
Given that our latent tokens lack inherent sequential order or spatial position information, conventional positional embedding approaches such as absolute positional encoding and relative positional encoding are not applicable. Instead, we associate each latent token with learnable positional tokens, which are added as residuals to the original latent tokens, enabling positional embedding adaptively updated during training.

\paragraph{Image Tokens}
For processing conditional images, we leverage complementary features from CLIP~\cite{clip21} and DINOv2~\cite{oquab2023dinov2}. The concatenated features serve as initial tokens for the MAR encoder, providing both semantic understanding and fine-grained pixel-level features. During training, we implement classifier-free guidance by randomly nullifying conditional input features with 0.1 probability, enhancing conditional generation quality~\cite{ho2022cfg}.

\paragraph{MAE Encoder and Decoder}
We concatenate images tokens with latent tokens and apply random masking with a ratio ranging from 0.7 to 1.0. Following the MAE architecture, we employ bidirectional attention mechanisms. The process involves first processing unmasked tokens through the MAE encoder, which is composed of multiple self-attention layers, and then concatenating the encoded tokens with mask tokens—using the same pre-defined learnable token for all masked positions. Learnable positional embeddings are incorporated into both masked and unmasked tokens before entering the MAE decoder, enabling position-aware token prediction.
\paragraph{Diffusion process}
The decoder generates each condition vector $\mathbf{z} \in \mathbb{R}^D$ for each token used in the diffusion process. A lightweight MLP-based denoising network then reconstructs ground truth tokens from Gaussian noise by optimizing:
\begin{equation}
\mathcal{L}(z,x) = \mathbb{E}{z,t}\left[|\epsilon - \epsilon\theta(x^t|t,z)|^2\right],
\end{equation}
where $z$ is the condition vector from MAE decoder and $x_t$ is the ground truth latent token provided by well-trained VAE. During inference, reverse diffusion process is applied to predict each set of tokens.

\begin{table*}
\centering
\begin{tabular}{l|ccc|ccc}
\hline
& \multicolumn{3}{c|}{GSO~\cite{gso22}} & \multicolumn{3}{c}{OmniObject3D~\cite{wu2023omniobject3d}} \\
Method & F-Score $\uparrow$ & CD $\downarrow$ & NC $\uparrow$ & F-Score $\uparrow$ & CD $\downarrow$ & NC$\uparrow$ \\
\hline
LGM~\cite{lgm24} & 0.745 & 0.813 & 0.685 & 0.738 & 0.821 & 0.677\\
CraftsMan~\cite{craftsman24}  & 0.776 & 0.785 & 0.687 & 0.771 & 0.798 & 0.675 \\
TripoSR~\cite{TripoSR2024}& 0.834  & 0.644 & 0.727 & 0.825 & 0.621 & 0.731\\
InstantMesh~\cite{xu2024instantmesh}& 0.923 & 0.415 & 0.780 & 0.918 & 0.427 &0.779  \\
Ours & \textbf{0.944}  & \textbf{0.351} & \textbf{0.835} & \textbf{0.931} & \textbf{0.364} & \textbf{0.826}  \\
\hline
\end{tabular}
\caption{\textbf{Comparison of different methods on GSO and OmniObject3D datasets.} Arrows ($\uparrow$/$\downarrow$) indicate whether higher or lower is better. Best results are in \textbf{bold}.}
\label{tab:campare}
\end{table*}
\subsubsection{MAR-HR}
We analyze the relationship between VAE latent token length and reconstruction error. While increasing token length improves VAE reconstruction quality, directly training MAR on longer sequences poses convergence challenges due to quadratic computational complexity. To achieve high-quality geometric details while maintaining computational efficiency, we implement a coarse-to-fine generation strategy using a super-resolution model (MAR-HR) that shares MAR-LR's architecture but generates high-resolution tokens conditioned on low-resolution latent tokens and image tokens. The training objective for MAR-HR is defined as:
\begin{equation}
\begin{aligned}
\mathcal{L}_{H}(z_{h},x_{h}) = \mathbb{E}{z_{h},t} \left[|\epsilon - \epsilon_\theta(x_{h}^{t} |t,z_{h})|^2\right],
\end{aligned}
\label{eq:sup}
\end{equation}
where $x_{h}$ represents high-resolution tokens from the VAE and $z_{h}$ denotes the condition vector from MAE decoder in our MAR-HR model.

\paragraph{Condition Augmentation}
To address the discrepancy between VAE-generated low-resolution tokens used during training and MAR-LR-generated tokens used during inference, we employ a condition augmentation strategy to mitigate compounding error~\cite{ho2021cascaded}. Our approach applies Gaussian noise to the low-resolution tokens $x_{l}$ through:
\begin{equation}
x_{l}{'} = t \epsilon+ (1-t)x_{l},
\end{equation}
where $\epsilon \sim \mathcal{N}(0, \mathbf{I})$ and $t \sim \mathcal{U}(0.4,0.6)$ during training, with $t$ fixed at 0.5 during inference. This augmentation is applied to both VAE-generated low-resolution tokens during training and MAR-LR-generated low-resolution tokens during inference before they are processed by MAR-HR.
This strategy effectively reduces compounding error and provides strong conditional guidance to MAR-HR, enabling faster convergence compared to direct high-resolution token training. The previous approach, as seen in~\cite{zhang2024clay}, requires substantial computational resources (hundreds of GPUs) and, as demonstrated in our experiments, yields poor convergence under the same limited GPU resources.

\subsection{Inference Schedules}
\paragraph{Generation Schedule}
During inference, we first generate a random token generation order. Then we extract image tokens from input image, which are fed into the MAE encoder-decoder architecture. From the decoder outputs, we select condition vector $z$ according to the predetermined generation order. Multiple tokens undergo parallel DDIM~\cite{song2022ddim} denoising processes simultaneously. The number of tokens $N_s$ processed in each auto-regressive step follows a cosine schedule as in~\cite{chang2022maskgit}, progressively increasing over $S$ total steps:
\begin{equation}
\centering
\begin{aligned}
N_{s} = \left\lfloor N(cos(\frac{s}{S }) -cos(\frac{s-1}{S})) \right\rfloor.
\end{aligned}
\label{eq:masking}
\end{equation}
This scheduling strategy is motivated by the observation that initial tokens are more challenging to predict, while later tokens become progressively easier to determine, similar to a completion task. Consequently, we generate fewer tokens in initial steps and gradually increase the number in later steps, rather than maintaining a constant generation rate across all steps. 
\paragraph{CFG Schedule}
We employ CFG in our diffusion model:
\begin{equation}
\centering
\begin{aligned}
\varepsilon = \varepsilon_{\theta}(x_t|t, z_u) + \omega_{s} \cdot (\varepsilon_{\theta}(x_t|t, z_c) - \varepsilon_{\theta}(x_t|t, z_u)),
\end{aligned}
\label{eq:cfg}
\end{equation}
where $z_c$ and $z_u$ are conditional and unconditional output from the MAE decoder, which serve as the condition for the diffusion model. We employ a linear strategy~\cite{mar24} for the CFG coefficient $\omega_{s}$, starting with lower values during the initial uncertain steps and progressively increasing it to $\lambda_{cfg}$. Specifically, in Eq.~\ref{eq:cfg}, we set $\omega_{s}=s*\lambda_{cfg}/S$.

\subsection{Discussion}
The MAR-3D architecture offers significant advantages over joint distribution modeling methods such as DiT by decomposing the complex joint distribution into temporal (diffusion) and spatial (auto-regressive) components. This decomposition, combined with our cascaded super resolution model, enables progressively increasing tokens. The effectiveness of this approach is comprehensively validated through ablation studies comparing against existing joint distribution modeling method by DiT~\cite{Peebles2022DiT}.

\section{Experiments}
\label{sec:exp}
\subsection{Implementation Details}
We train our pyramid VAE using a hierarchical point cloud representation with 16384, 4096, and 1024 points from the highest to lowest level, along with 20480 sampled ground truth occupancy values. For cascaded MAR model training, we use the conditional images and latent tokens sampled from the pyramid VAE, with batch sizes of 32 and 8 per GPU for MAR-LR and MAR-HR respectively. Training and inference details are provided in the supplementary.
\subsection{Training Strategies}
\paragraph{Training Data Curation}
Our training procedure utilizes carefully curated data from Objaverse~\cite{wu2023omniobject3d}, implementing a two-stage approach.We first train on 260K geometrically diverse meshes with partial low texture quality for 200 epochs, followed by fine-tuning on 60K high-quality meshes with natural textures for 100 epochs.
\paragraph{Rotation Augmentation}
For each 3D mesh, we render 56 conditional views using a structured rendering approach.  We first uniformly sample 8 base views with azimuth angles ranging from 0° to 360°. Each base view is then augmented with 6 different random rotations. To ensure consistency between the input images and 3D latent tokens, we apply corresponding rotations to the 3D meshes before VAE encoding. This strategy effectively avoids the ambiguity that would arise from mapping the same canonical mesh to different conditional view images.
\begin{figure*}
\centering 
\includegraphics[width=0.92\textwidth]{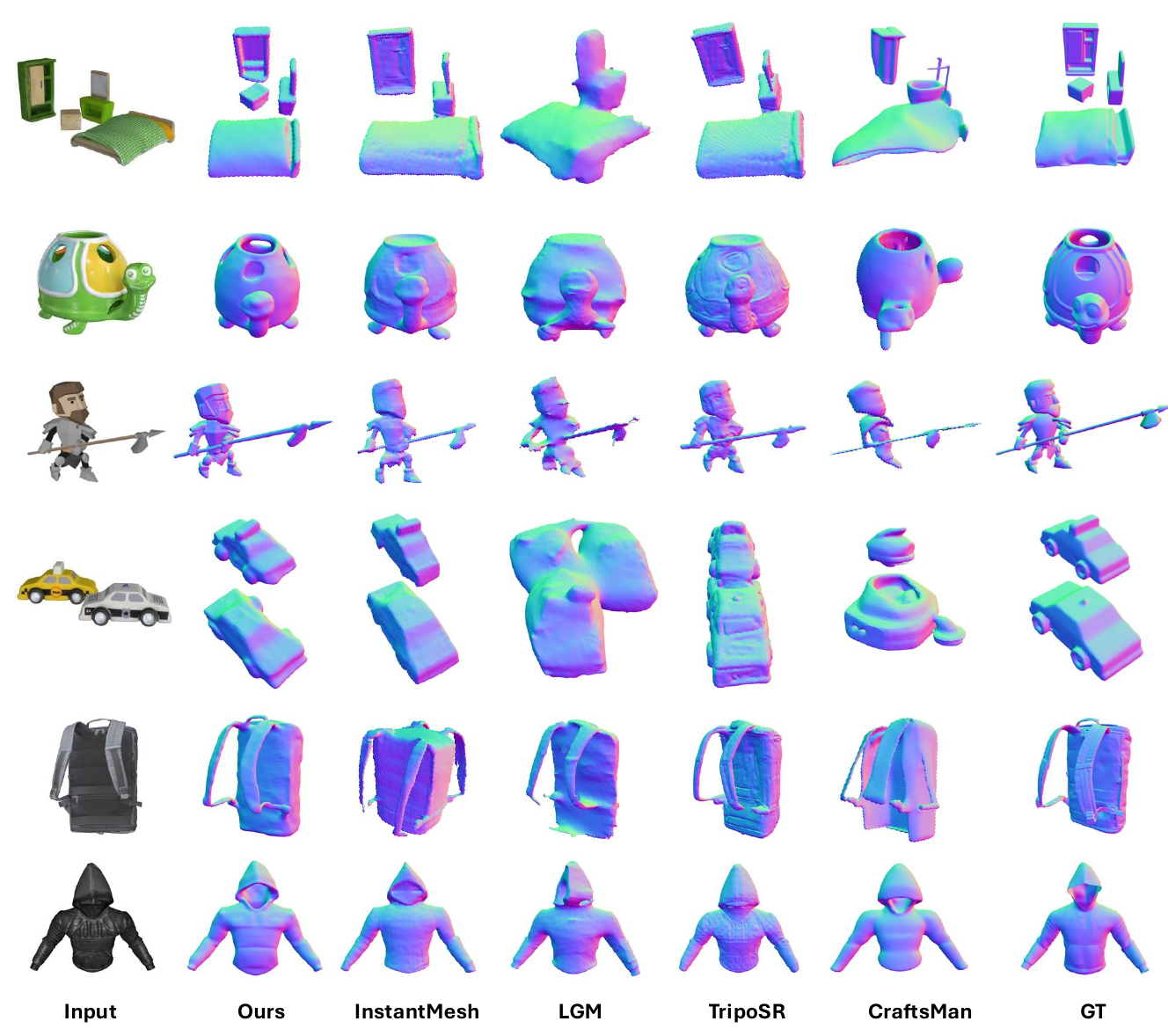} 
\caption{\textbf{Comparison on rendered normal map.} We visualize the normal map rendered by our method and other baseline methods.}
\label{fig:campare}
\end{figure*}

\subsection{Comparison Results}
\paragraph{Evaluation Settings}
We compare our method with recent single-view approaches: InstantMesh~\cite{xu2024instantmesh} (multiview diffusion with feed-forward SDF), LGM~\cite{lgm24} (feed-forward 3D Gaussians), TripoSR~\cite{TripoSR2024} (large reconstruction model without diffusion), and CraftsMan~\cite{craftsman24} (VAE-diffusion pipeline similar to ours). All evaluations use official pre-trained models.
We evaluate on GSO~\cite{gso22} and OmniObject3D~\cite{wu2023omniobject3d} datasets, which contain unseen real-scanned objects. To ensure meaningful comparison, we first remove too simple categories such as boxes and then randomly sample 100 shapes from each dataset (200 total). We assess performance using F-score, Chamfer distance (CD), and Normal Consistency (NC) between predicted and ground truth meshes, after normalization and ICP alignment.
\paragraph{Quantitative Evaluation}
The quantitative comparison with existing single-view mesh reconstruction and generation methods is presented in Tab.~\ref{tab:campare}. Our method significantly outperforms baseline approaches across all evaluation metrics on both test datasets, demonstrating strong generalization to unseen data. On the GSO dataset, our method achieves a Chamfer Distance (CD) of 0.351, reducing the geometric error by 15.4\% compared to InstantMesh (0.415) and showing even more substantial improvements over naive diffusion methods like CraftsMan (0.785). This significant reduction in CD indicates more accurate geometric reconstructions. The improvement is consistent on the OmniObject3D dataset, where our method maintains a low CD of 0.364, outperforming InstantMesh (0.427) and other competitors by a large margin. Additionally, we achieve strong performance in other metrics, with F-Score reaching 0.944 and Normal Consistency (NC) of 0.835 on GSO, further validating the effectiveness of our approach.
\paragraph{Qualitative Evaluation}
We conduct qualitative evaluation in Fig.~\ref{fig:campare} and showcase representative examples demonstrating our method's capabilities in handling challenging cases, including multi-object scenes, intricate geometric structures, and meshes with topological holes.
We visualize the normal map rendered from the meshes generated by these methods. 
As demonstrated in the figure, our method exhibits superior reconstruction capabilities compared to existing approaches. LGM struggles with geometric accuracy due to multi-view inconsistency and challenges in converting 3D Gaussians to high-quality meshes. While CraftsMan employs a similar VAE-diffusion pipeline, its naive design leads to incorrect geometry and incomplete reconstructions. InstantMesh produces relatively high-quality meshes but faces challenges with multi-object scenes and occasionally generates geometries inconsistent with input images, such limitation also observed in TripoSR without using any generative prior. Our method achieves better geometric fidelity and completeness through careful design of both the VAE architecture and generation pipeline.
\subsection{Ablation Study}
\begin{figure}
\centering 
\includegraphics[width=0.45\textwidth]{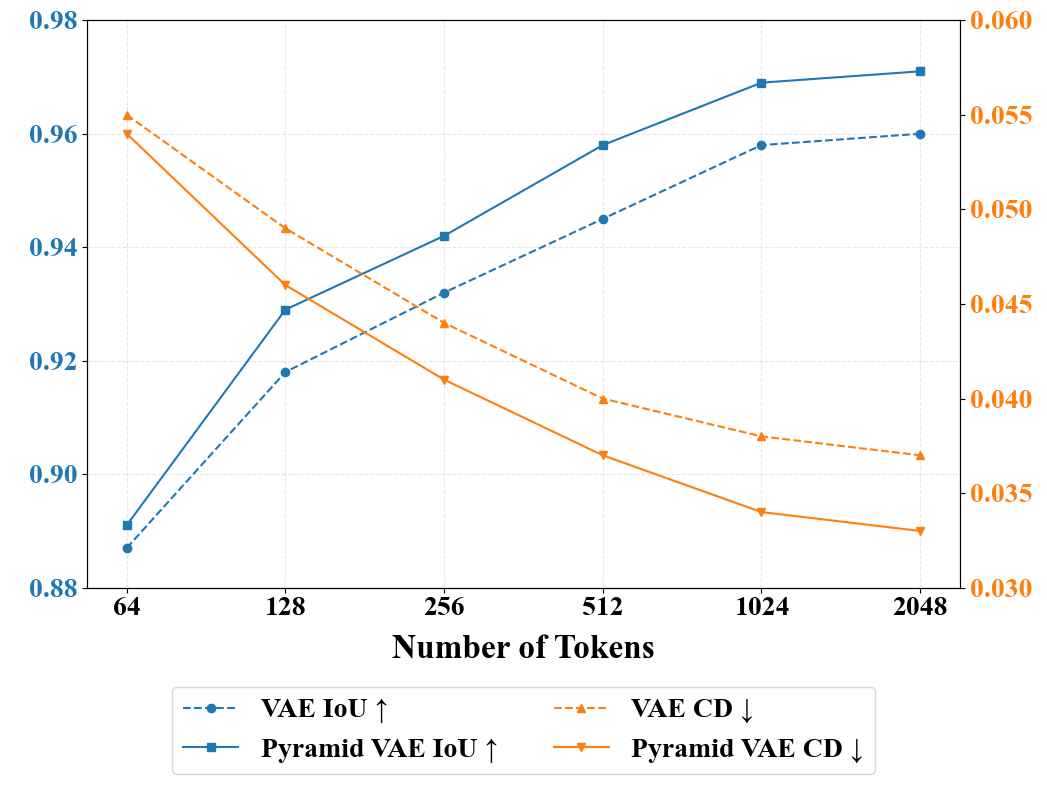} 
\caption{\textbf{VAE Metrics with varying number of tokens.} We show the reconstructed mesh CD and IoU of our Pyramid VAE vs the original VAE in terms of different number of tokens.}
\label{fig:vae_mul}
\end{figure}
\begin{figure}
\centering 
\includegraphics[width=0.45\textwidth]{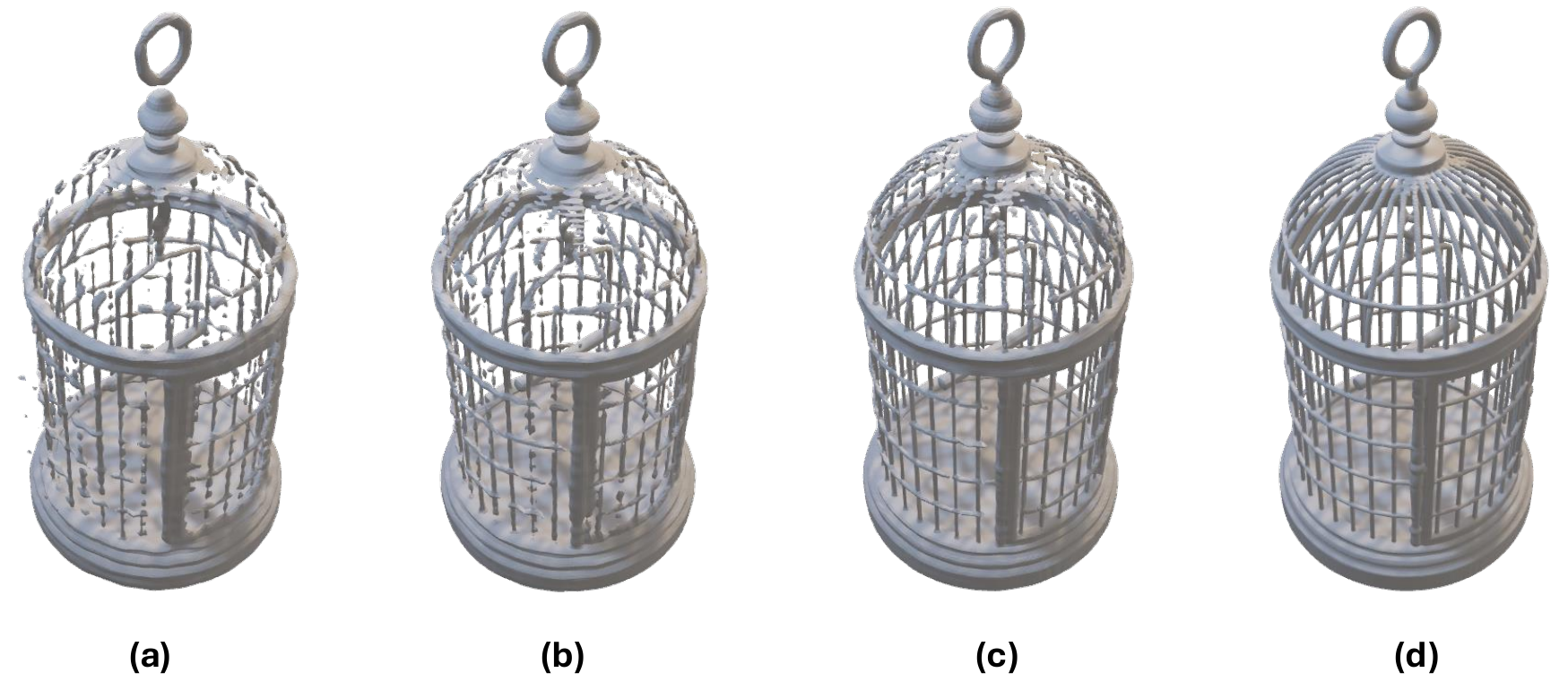} 
\caption{\textbf{Visual comparison of VAE reconstruction.} (a)-(c) show reconstruction results from single-level VAE compressed with 256, 1024, and 2048 latent tokens respectively. (d) demonstrates the result from our Pyramid VAE using 1024 tokens.}
\label{fig:mul}
\end{figure}

\paragraph{VAE Ablations}
We evaluate the VAE performance by analyzing latent token length and our multi-resolution pyramid design. The quantitative results in Fig.~\ref{fig:vae_mul} demonstrate that reconstruction error decreases with increasing token length, though improvements beyond 1024 tokens become marginal. Our pyramid design consistently enhances reconstruction quality, with 1024 tokens under Pyramid VAE outperforming 2048 tokens without it. Based on this analysis, we select 256 tokens for MAR-LR and 1024 tokens for MAR-HR, striking an optimal balance between computational efficiency and generation quality. As shown in Fig.~\ref{fig:mul}, more geometric details are shown in our Pyramid VAE (d) using less tokens compared with single-level VAE (c).

\begin{table}
\centering
\begin{tabular}{l|ccc}
\hline
Setting &  F-Score $\uparrow$ & CD $\downarrow$ &NC$\uparrow$  \\
\hline
 w/o Pyramid VAE & 0.928 & 0.397 & 0.807 \\
w/o condition aug& 0.902 & 0.435 & 0.789 \\
 w/o MAR-HR & 0.921 & 0.411& 0.794 \\
 w/o rotation aug& 0.934 & 0.369 & 0.821 \\
\hline
full & 0.944 & 0.351 & 0.835 \\
\hline
\end{tabular}
\caption{\textbf{Ablation study of different components in our method.} $\uparrow$ indicates higher is better, and $\downarrow$ indicates lower is better.}
\label{tab:ablation_t}
\end{table}
\begin{figure}
\centering 
\includegraphics[width=0.48\textwidth]{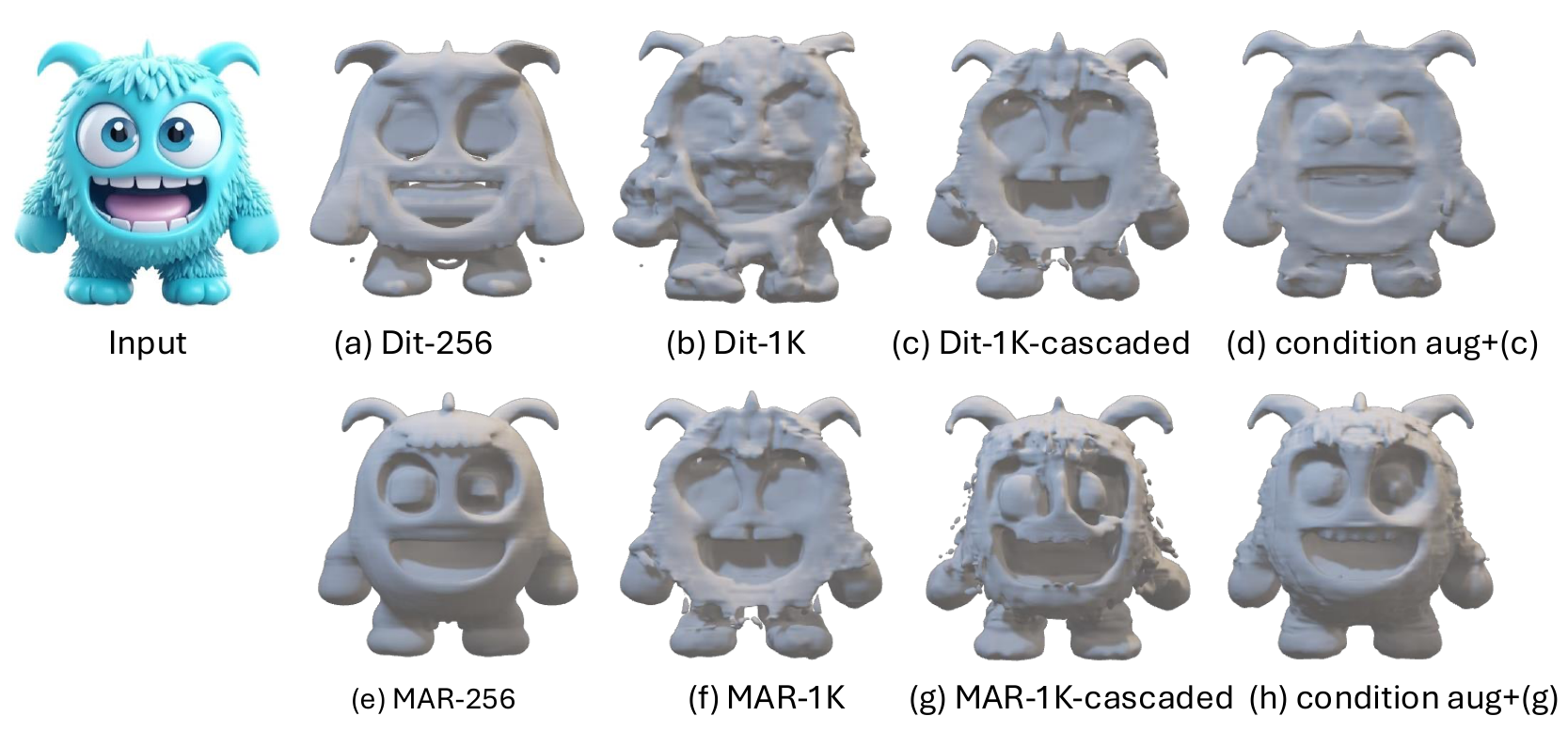} 
\caption{\textbf{Ablation study on token resolution and model scaling strategies.} Results (a)-(h) demonstrate different model configurations and settings, with detailed analysis provided in the main text.}
\label{fig:abl}
\end{figure}

\paragraph{Generation Ablations}
Tab.~\ref{tab:ablation_t} presents our ablation study for generation on several key components on GSO datset: Pyramid VAE, condition augmentation, MAR-HR, rotation augmentation. The results demonstrate that our Pyramid VAE enhances generation quality through its improved latent space, adding MAR-HR with condition augmentation significantly improves cascaded generation quality by mitigating compounding error, while view augmentation reduces view ambiguity. 
We also evaluate our cascaded MAR design through comparison with a DiT implementation, as illustrated in Fig.~\ref{fig:abl}. Our base model with 256 latent tokens (e) achieves superior geometry quality compared to the DiT version with the same latent tokens (a). When directly increasing to 1024 tokens, both our model (f) and DiT (b) show degraded performance due to convergence issues. While the cascaded model enhances detailed generation, error propagation from the low-resolution model introduces significant noise in both MAR (g) and DiT version (c). Our MAR-HR with condition augmentation (h) successfully up-scales token resolution and achieves detailed generation, demonstrating clear advantages over the DiT version (d). This ablation study demonstrates that our cascaded MAR with condition augmentation offers an effective and efficient solution for scaling up the token resolution. Quantitative comparison are provided in the supplementary.
\section{Conclusion}
We present a new 3D generation paradigm that combines auto-regressive and diffusion models while addressing key challenges in scaling to longer tokens. Through a Pyramid VAE and cascaded training with condition augmentation strategy, we progressively refine low-resolution tokens into high-resolution ones. Both quantitative and qualitative results demonstrate the effectiveness of our method, highlighting the potential of auto-regressive 3D generation.
\label{sec:conclusion}

\paragraph{Acknowledgement.}
This research / project is supported by the National Research Foundation (NRF)
Singapore, under its NRF-Investigatorship Programme (Award ID. NRF-NRFI09-0008).

{
\small
\bibliographystyle{ieeenat_fullname}
\bibliography{main}
}



\end{document}